\documentclass[letterpaper, 10 pt, conference]{ieeeconf}  

\IEEEoverridecommandlockouts                              

\usepackage{amsmath} 
\usepackage{amssymb}  
\usepackage{float}
\usepackage{booktabs}
\usepackage{multirow}
\usepackage{cite}
\usepackage{graphicx}
\usepackage{hyperref}
\usepackage{makecell}
\usepackage[utf8]{inputenc}
\usepackage{url}
\usepackage{mathtools}
\usepackage{gensymb}
\usepackage{color}
\usepackage{subfig}
\usepackage{adjustbox} 
\usepackage{booktabs}
\usepackage{subcaption} 
\usepackage{arydshln}

\title{\LARGE \bf
XRDSLAM: A Flexible and Modular Framework for Deep Learning based SLAM
}

\author{Xiaomeng Wang$^{1*}$ \quad Nan Wang$^{1*}$ \quad Guofeng Zhang$^{2\dagger}$
\\$^{1}$SenseTime Research \quad  $^{2}$State Key Lab of CAD\&CG, Zhejiang University
\thanks{$^*$ Equal Contribution}
\thanks{$^\dagger$ Corresponding Author}
}

\begin{document}

\maketitle
\thispagestyle{empty}
\pagestyle{empty}

\begin{abstract}
In this paper, we propose a flexible SLAM framework — XRDSLAM. It adopts a modular code design and a multi-process running mechanism, providing highly reusable foundational modules such as unified dataset management, 3d visualization, algorithm configuration, and metrics evaluation. It can help developers quickly build a complete SLAM system, flexibly combine different algorithm modules, and conduct standardized benchmarking for accuracy and efficiency comparison. Within this framework, we integrate several state-of-the-art SLAM algorithms with different types, including NeRF and 3DGS based SLAM, and even odometry or reconstruction algorithms, which demonstrates the flexibility and extensibility. We also conduct a comprehensive comparison and evaluation of these integrated algorithms, analyzing the characteristics of each. Finally, we contribute all the code, configuration and data to the open-source community, which aims to promote the widespread research and development of SLAM technology within the open-source ecosystem.
    
\begin{itemize}
    \item Open Source:
    \href{https://github.com/openxrlab/xrdslam}{https://github.com/openxrlab/xrdslam}
\end{itemize}

\end{abstract}

\section{INTRODUCTION}

SLAM (Simultaneous Localization and Mapping) technology plays a crucial role in fields such as autonomous driving, robotics, and AR/VR. By analyzing and fusing sensor data, it achieves accurate 6DoF (Degree of Freedom) localization of devices and constructs environmental maps in real time, providing foundational capabilities for applications like autonomous driving, robot navigation, and AR/VR. 
The form of map representation is very crucial to the design of SLAM algorithms. 
While traditional SLAM algorithms\cite{Mur:2015:ORB,campos2021orb} often express scenes based on sparse or semi-dense point clouds and manage maps through hand-written rules, which limits the development of SLAM technology.

In recent years, technologies like Neural Radiance Fields (NeRF)\cite{mildenhall2021nerf} and 3D Gaussian Splatting (3DGS)\cite{3dgs} have made significant progress in novel view synthesis. Learning based scene representations leverage advantages such as differentiable rasterization, continuous modeling, noise handling, and hole-filling, injecting new energy into the SLAM field\cite{tosi2024nerfs}. Some recent works have attempted to incorporate NeRF and 3DGS into SLAM systems. Some methods\cite{yang2022vox,zhu2022nice} directly use RGB-D data as color and depth supervision to train network parameters and optimize poses. Other methods\cite{zhu2024nicer} rely solely on RGB data, optimizing networks and poses by adding regularization constraints. Some efforts also combine these new technologies with traditional SLAM methods by introducing techniques like ICP\cite{zhang2021iterative} and ORB-SLAM\cite{Mur:2015:ORB,campos2021orb}. Additionally, some work attempts to integrate extra networks to provide prior information such as depth priors, optical flow, and normals, aiming to improve training speed and reconstruction quality.


However, with the continuous emergence of new papers and technologies, tracking progress and integrating code becomes increasingly challenging. Many research implementations are scattered across independent code repositories, lacking a unified development and evaluation process. This makes fair and comprehensive comparisons become very difficult. Additionally, the complexity of SLAM systems raises high technical barriers and development costs for building a complete SLAM system from scratch. Inspired by the success of NeRFStudio\cite{tancik2023nerfstudio} in the NeRF domain, we introduce XRDSLAM, a scalable and multifunctional framework for deep learning based SLAM. Our main contributions in this work are as follows:

\textbf{Modular Design:} XRDSLAM adopts a multi-process mechanism and modular code design, decoupling the tracking, mapping, and visualization processes. The framework provides modular data input/output, configuration parsing, visualization, and result export function modules, allowing users to flexibly combine and replace different algorithm modules to achieve rapid iteration and optimization.

\textbf{Unified Pipeline:} 
The framework offers a unified SLAM development process with reusable components, allowing developers to easily create a full SLAM algorithm. It also standardizes data handling and evaluation, enables fair and effective evaluations between different SLAM algorithms. Specifically, we have designed a dedicated \textit{Model} class to manage all operations of deep learning models,  making it easier for developers to build learning-based algorithms.

\textbf{Integration of SOTA Algorithms:} We integrate state-of-the-art (SOTA) SLAM algorithms, which validates the framework's flexibility and extensibility, and also facilitates the open-source community's learning and comparison of different algorithms.

\textbf{Open-sourced Code:} All of the code, configuration and data are publicly available, aiming to promote community-driven development and the widespread application of these technologies. This work is part of OpenXRLab \footnote{\url{https://github.com/openxrlab}}.

\section{Related Work}
In this section, we briefly review learning based SLAM systems, as well as frameworks and platforms.

\subsection{Learning based SLAM}

Benefit from the rapid development of deep learning technology, the research paradigm of SLAM is also continuously evolving. Early methods mainly involved introducing mature work of deep learning, such as feature learning\cite{teed2020raft,teed2021droid}, depth estimation\cite{tateno2017cnn}, and semantic segmentation\cite{yu2018ds}, to enhance the performance of SLAM. In recent years, with the emergence of NeRF and 3DGS technologies, the research focus in the SLAM field has shifted towards innovation in map representation.

iMAP\cite{sucar2021imap} is the first to use NeRF in dense visual SLAM, using an MLP for continuous scene modeling. However, it struggled with large scenes and real-time reconstruction. To improve this, later methods introduced multi-resolution grids\cite{zhu2022nice}, dynamic octrees\cite{yang2022vox}, and multi-scale feature planes\cite{johari2023eslam}, improving scalability and scene understanding. Techniques like Point-SLAM\cite{sandstrom2023point} and Loopy-SLAM\cite{liso2024loopy} used neural point clouds, avoiding grid resolution issues and improving detail and surface reconstruction.

3DGS-based SLAM \cite{keetha2024splatam} systems have emerged, leveraging Gaussian shape regularization\cite{matsuki2023gaussian}, densification and pruning of Gaussian point clouds\cite{yan2024gs}, submap management\cite{yugay2023gaussian}, and uncertainty awareness\cite{hu2024cg} to enhance accuracy and efficiency.


Early neural rendering SLAM methods\cite{yang2022vox,zhu2022nice,keetha2024splatam} relied on RGBD sensor data for depth supervision. More recent approaches\cite{zhu2024nicer,zhang2024glorie} have explored using only RGB images for dense SLAM. In monocular settings, optimization can be slower due to fewer constraints, but some methods\cite{zhu2024nicer,belos2022mod,naumann2024nerf,zhu2024mgs} add extra constraints like depth, normal vectors, and optical flow to improve performance. Recent methods, such as Glorie-SLAM\cite{zhang2024glorie}, SplaSLAM\cite{sandstrom2024splat} and DROID-SLAM\cite{teed2021droid}, use dense optical flow and multi-view consistency checks to produce reliable depth maps, significantly boosting SLAM performance.

\begin{figure*}[ht]
    \centering
    \includegraphics[width=\linewidth]{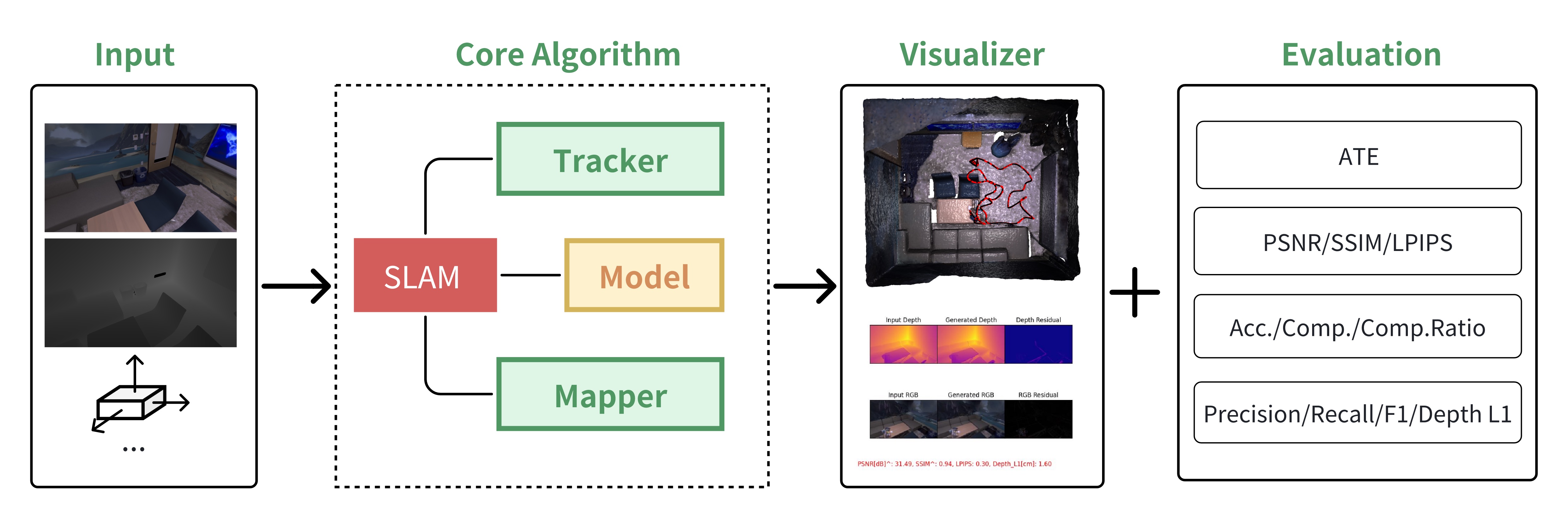}
    \caption{System Overview of XRDSLAM. It takes RGB, depth, IMU, and other data as \textit{Input}, processed through the \textit{Tracker} and \textit{Mapper} processes, with core SLAM algorithm modules providing essential functionality to these processes. A reusable  \textit{Visualizer} module is used to display the algorithm's outputs, and an \textit{Evaluation} module is included for comprehensive metric assessment. The section within the dashed box needs to be customized by developers.}
    \label{fig:XRDSLAM}
\end{figure*}

\subsection{Frameworks of SLAM}

In traditional SLAM, systems like ORB-SLAM\cite{campos2021orb}, VINS-Fusion\cite{qin2018vins,qin2019general}, and OpenVINS\cite{geneva2020openvins} are highly influential. They support various sensors (monocular, stereo, RGBD) and offer key functionalities such as loop detection and relocalization, providing SOTA localization accuracy. OpenVINS is particularly notable for its detailed documentation, which helps researchers and engineers with development and research. GSLAM\cite{zhao2019gslam} offers a comprehensive cross-platform SLAM solution, including evaluation and visualization tools, advancing traditional SLAM applications.

NeRFStudio\cite{tancik2023nerfstudio} is a modular framework that simplifies the process of creating, training, and testing NeRF models. SDFStudio\cite{Yu2022SDFStudio} builds on NeRFStudio, offering multiple map representations and a unified framework for neural implicit surface reconstruction.

In the NeRF and 3DGS SLAM domain, open-source algorithms like NICE-SLAM\cite{zhu2022nice}, Vox-Fusion\cite{yang2022vox}, and SplaTAM\cite{keetha2024splatam} are gaining attention. These methods vary in aspects such as data processing, keyframe selection, and optimization parameters, making it challenging to compare their performance. A unified RGBD-SLAM\cite{hua2024benchmarking} benchmark framework has been proposed to evaluate the impact of different NeRF representations and geometric rendering on SLAM systems. However, there is still a need for a unified framework to streamline data input, visualization, and evaluation for easier integration of new algorithms.

\section{System}

\subsection{Framework Overview}
XRDSLAM provides a highly configurable, efficient, and easily extensible framework. Inspired by NeRFStudio, XRDSLAM adopts a modular code design, organizing SLAM components through inheritance and polymorphism. It handles parameters through a centrally managed configuration class and accelerates tracking, mapping, and visualization tasks using multi-process parallel processing, significantly enhancing system performance. XRDSLAM combines flexible configuration, resource sharing, and a multi-process mechanism to create a versatile and efficient development platform.

The XRDSLAM framework, as shown in Fig. \ref{fig:XRDSLAM}, mainly handles the interaction of four parts:
\subsubsection{\textbf{Dataset Input}} The framework needs to receive data from one or multiple sensors, such as RGB images, depth images, IMU data, and etc. XRDSLAM processes different formats of datasets through multiple \textit{Dataset} classes, providing a unified data loading interface and preprocessing functions.
\subsubsection{\textbf{Core SLAM Algorithms}} XRDSLAM organizes the core SLAM algorithm through the \textit{Algorithm} and \textit{Model} classes. Users can quickly implement and optimize the SLAM algorithm by inheriting and replacing components with them. And the \textit{Model} class is invoked by the \textit{Algorithm} class. Developers can design and manage deep learning algorithms through the \textit{Model} class.
\subsubsection{\textbf{Visualization}} XRDSLAM provides online and offline visualization tools. Online visualization tools display the status of the SLAM system in real-time, including 2D images rendering and 3D scene representations, assisting users in debugging and optimization. Offline visualization tools are used for rendering and displaying the results of the SLAM algorithm, supporting saving as video files for subsequent analysis.
\subsubsection{\textbf{Evaluation}} XRDSLAM automatically saves the results of the SLAM system and uses the \textit{Evaluation} module to assess the trajectory and reconstruction results.

\subsection{Multipie Processes} 
SLAM systems typically consist of multiple modules, such as tracking, mapping, and loop closure detection. To achieve module decoupling and independent operation, XRDSLAM employs a multi-process mechanism. The framework includes three key subprocesses: \textit{Tracker}, \textit{Mapper}, and \textit{Visualizer}.

\subsubsection{\textbf{Tracker}} It is responsible for processing input data and outputting camera pose information. It receives data from the \textit{Dataset} class, calculates the camera's pose information, and passes the results to the \textit{Mapper} process.

\subsubsection{\textbf{Mapper}} It extracts keyframes and optimizes pose and map information.  The \textit{Mapper} process receives frame data and pose parameters from the \textit{Tracker} process to continuously expand the map, further optimize the camera pose and map parameters, and handle tasks such as map fusion and updates.

\subsubsection{\textbf{Visualizer}} It's responsible for visualizing the trajectory , rendered images, and reconstruction results. It receives data from the \textit{Tracker}  process and visualizes them in real-time.

Through PyTorch's\cite{paszke2019pytorch} \textit{multiprocessing} module, XRDSLAM has implemented the creation and initiation of these processes, achieving parallel processing. This multi-process architecture not only enhances the system's efficiency and real-time performance but also allows each module to independently manage resources, reducing resource competition and conflicts, thereby strengthening the system's stability and reliability. Moreover, it also supports independent expansion and optimization of modules, further enhancing the framework's flexibility.

\subsection{Data sharing mechanism}

The XRDSLAM framework achieves inter-process data sharing and coordination through the following methods:

\subsubsection{\textbf{BaseManager}} It is used to create shared instances of the \textit{Algorithm} class, allowing the \textit{Tracker} and \textit{Mapper} processes to conveniently share data and functionalities. By centrally managing core tracking and mapping functionalities, it facilitates efficient data exchange and functional collaboration.

\subsubsection{\textbf{Queue}} It is used for passing data streams between processes. The \textit{Tracker} process packages the processed pose information and input frame information into a \textit{Frame}, and passes it to the \textit{Mapper} process through the $map\_buffer$. It also passes system status information to the Visualizer process through the $viz\_buffer$, achieving efficient data stream transmission.

\subsubsection{\textbf{Event}} It is used for operation and status synchronization between processes. The events $event\_ready$ and $event\_processed$ are used to indicate the readiness and completion of data processing, ensuring effective coordination and synchronization between the \textit{Tracker} and \textit{Mapper} processes.

Through these mechanisms, XRDSLAM is able to maintain consistency in data and algorithmic states across different processes, while effectively managing data flow and coordination between processes using buffers and event synchronization.

\section{SLAM Components}

XRDSLAM adopts a modular design, allowing each SLAM component to be independently configured and expanded, thereby enhancing the system's efficiency and maintainability. The entire system includes core algorithm components and general components. To reduce redundant development, evaluation, and visualization of general modules are abstracted into utility classes.

\begin{figure}
    \centering
    \includegraphics[width=\linewidth]{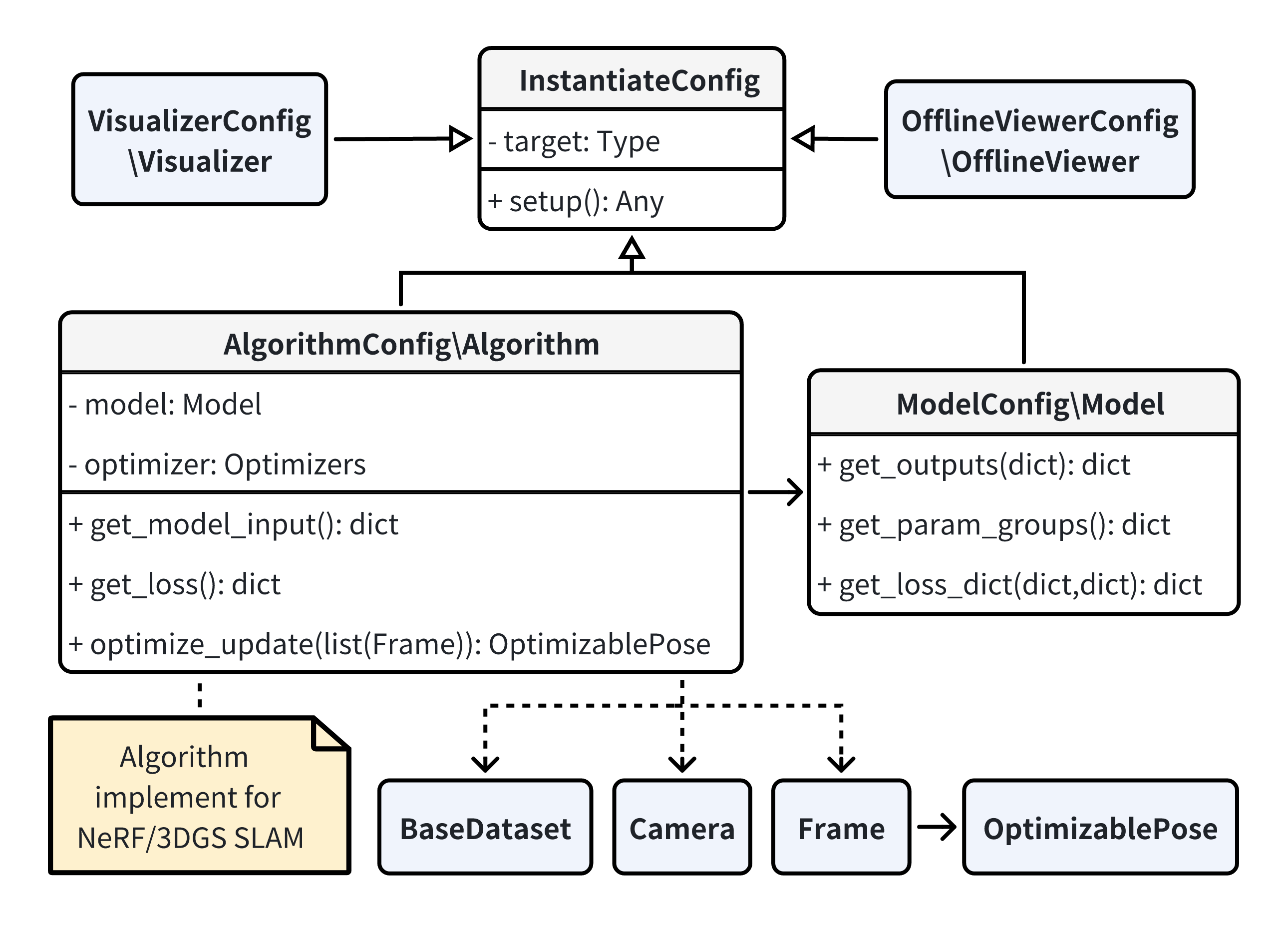}
    \caption{Components of XRDSLAM.}
    \label{fig:SLAM_components}
\end{figure}
\subsection{Basic Classes}
Basic components include commonly used base classes for functions such as configuring parameters, data reading, pose transformation, and camera model.

\subsubsection{\textbf{Config}}
Like NeRFStudio, XRDSLAM uses configuration classes to organize and manage the parameters and components of the SLAM algorithm. The configuration class inherits from \textit{InstantiateConfig} and dynamically instantiates the corresponding objects through the \textit{setup} method. Each component in the system can be configured and instantiated in this unified manner. \textit{XRDSLAMerConfig} centrally manages all parameters related to SLAM operation, and through this configuration class, users can conveniently control the behavior of the SLAM system, simplifying the parameter management and debugging process. The use of configuration classes not only improves the readability of the code but also enhances the flexibility and extensibility of the system.

\subsubsection{\textbf{BaseDataset}}

Based on the NICE-SLAM\cite{zhu2022nice} code, XRDSLAM supports multiple types of datasets through the \textit{get\_dataset} method and allows easy expansion or switching of data sources through configuration to meet different data needs. The dataset loading is implemented by multiple dataset classes and helper functions that inherit from \textit{BaseDataset}. These utility classes are compatible with various data formats and sources, providing a unified camera model and data loading interface, which facilitates integration into the SLAM framework. Currently supported datasets include Replica\cite{straub2019replica}, ScanNet\cite{dai2017scannet}, TUM RGBD\cite{sturm2012benchmark}, 7-Scenes\cite{glocker2013real}, Co-Fusion\cite{runz2017co}, and Euroc\cite{burri2016euroc},  with main functions including reading RGB and depth images, IMU data, and camera poses, as well as performing necessary preprocessing (such as undistortion, downsampling).

\subsubsection{\textbf{OptimizablePose}}

The \textit{OptimizablePose} class is used for pose optimization in SLAM, handling rotations and translations.  We extended the \textit{OptimizablePose} class from the Vox-Fusion\cite{yang2022vox} code to support two rotation representation methods: axis-angle and quaternions. It utilizes PyTorch’s automatic differentiation feature for gradient optimization, efficiently converting between pose parameters and rotation matrices. The \textit{OptimizablePose} class is included within the \textit{Frame} class.

\subsubsection{\textbf{Frame}}

The \textit{Frame} class is used to manage single-frame data in the SLAM system, including RGB images, depth images, and poses. It encapsulates this information and provides methods for setting, getting, and optimizing. By inheriting from \textit{torch.nn.Module}, it facilitates management and optimization within the deep learning framework.

\subsubsection{\textbf{Camera}}

The \textit{Camera} is used for managing the camera model in XRDSLAM. Currently, the supported model is the pinhole model. The camera's intrinsic parameters and resolution information are stored in the \textit{Camera} class.


\subsection{Core Modules}

\subsubsection{\textbf{Algorithm}}

The \textit{Algorithm} class is an abstract base class that defines the core interfaces and methods required to implement SLAM algorithms. It provides features such as selecting optimized frames, obtaining model inputs, computing losses, preprocessing, and postprocessing, constructing a general iterative optimization process for the SLAM system. Additionally, the class includes abstract methods for optimizer configuration, system state maintenance, and updates.

Different SLAM algorithms can be implemented by inheriting from the \textit{Algorithm} class to achieve custom functionality to meet specific needs. The XRDSLAM framework has already integrated various algorithms, such as NICE-SLAM, Co-SLAM\cite{wang2023co}, Vox-Fusion, Point-SLAM, and SplaTAM. Each algorithm extends specific optimization and processing procedures based on inheriting from the \textit{Algorithm} class, demonstrating the framework's flexibility and extensibility.

Furthermore, the \textit{Algorithm} class decouples the tracking and mapping processes, allowing users to replace and combine different tracking and mapping methods to quickly verify various solutions. 

Specifically, when the tracking function directly uses the input ground-truth poses, the system can transition from traditional SLAM modes to incremental reconstruction processes. NeuralRecon\cite{sun2021neuralrecon} is integrated into XRDSLAM through this approach.
 For algorithms that do not require mapping, such as DPVO\cite{teed2024deep}, developers can skip the \textit{Mapper} module and implement only the pose tracking functionality in the \textit{Tracker} module.

\subsubsection{\textbf{Model}}

The \textit{Model} class is the base class for all specific network models, providing the basic structure and general interfaces for the models. It inherits from PyTorch's \textit{nn.Module} and is used to define and manage the main components of deep learning models. The core responsibilities of the \textit{Model} class include forward propagation, loss calculation, and obtaining optimized parameters.

Subclasses are configured based on the specific network structure of the model, including encoders, decoders, etc., managing model inputs and outputs, loss calculations, map updates, and processes such as NeRF and 3DGS rendering. These subclasses implement the management and updating of different map representations, such as multi-resolution feature grids, sparse voxel octrees, hash grids, neural point clouds, 3DGS, etc.

\begin{table*}[h]
    \centering
    \caption{We integrate different types of SLAM, including NeRF / 3DGS-based SLAM, odometry and reconstruction, covering different scene encodings\cite{tosi2024nerfs} and geometric representations. Bolded fonts mean the better performance amoung SLAM or reconstruction.}
    \begin{tabular}{cccccccc}
        \toprule
 \textbf{Type}& \multicolumn{4}{c}{NeRF based SLAM}& 3DGS based SLAM& Odometry&Reconstruction\\
        \midrule
         \textbf{Algorithm}&  NICE-SLAM&  Co-SLAM&  Vox-Fusion&  Point-SLAM&  SplaTAM&  DPVO& NeuralRecon\\
         \hline
         \textbf{Scene Encoding}
&  \makecell[c]{Hierarch Grid\\ + MLP}
&  \makecell[c]{Hash Grid\\ + MLP}
&  \makecell[c]{Octree Grid\\ + MLP}
&  \makecell[c]{Neural Points\\ + MLP}
&  3D Gaussians
&  -& \makecell[c]{Hierarch Grid\\ + MLP}
\\
\hdashline[1pt/3pt]
         \textbf{Geometry}&  Occupancy&  SDF&  SDF&  Occupancy&  Density&  Sparse Points
& SDF\\
\hline
 \textbf{GPU Mem.(GB)}$\downarrow$& 3.12& \textbf{2.93} & 10.72& 8.25 & 4.01& 3.75&10.10\\
\hdashline[1pt/3pt]
  \textbf{FPS}$\uparrow$& 0.53 & \textbf{1.31} & 0.38 & 0.09 & 0.18 & 2.76 &\textbf{\textit{3.55}} \\
\hdashline[1pt/3pt]
         {\textbf{Visualization}}& 
         \includegraphics[width=0.06\textwidth]{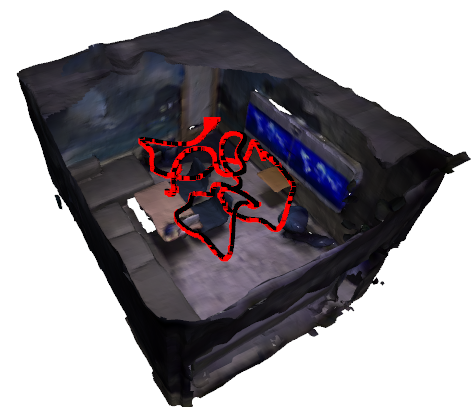} &  \includegraphics[width=0.06\textwidth]{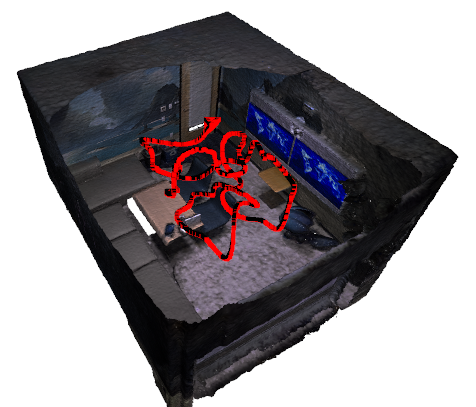} &  \includegraphics[width=0.06\textwidth]{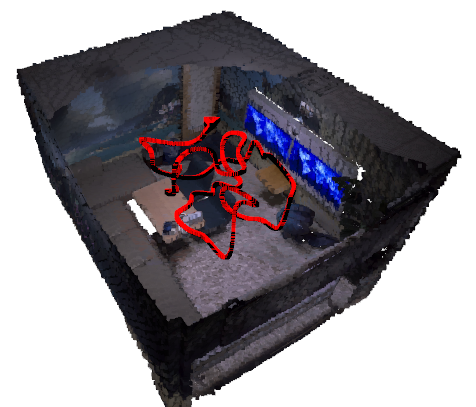} &  \includegraphics[width=0.06\textwidth]{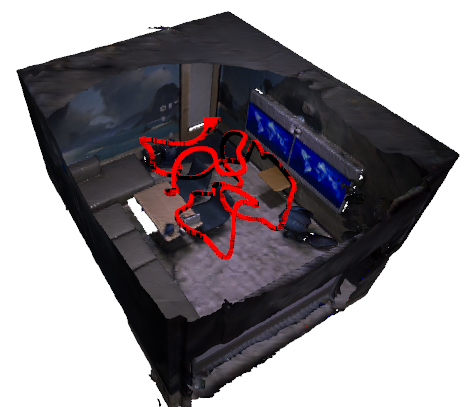} &   \includegraphics[width=0.06\textwidth]{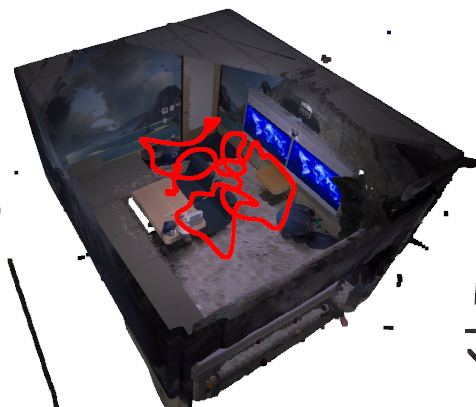}&   \includegraphics[width=0.06\textwidth]{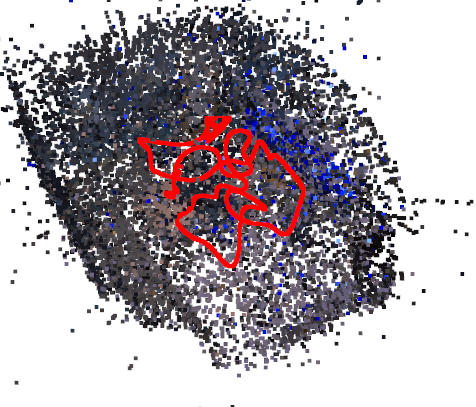}&  \includegraphics[width=0.06\textwidth]{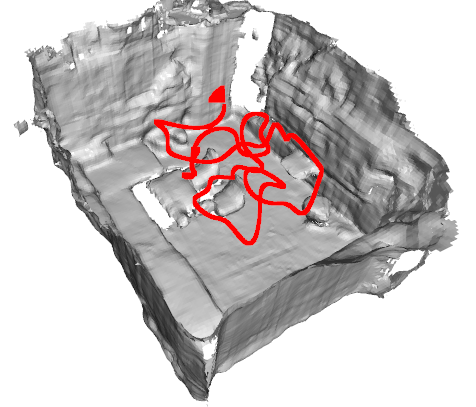}\\
         \bottomrule
    \end{tabular}
    \label{tab:SLAM_implements}
\end{table*}

\subsubsection{\textbf{Visualizer and OfflineViewer}}

The \textit{Visualizer} class is responsible for the real-time visualization of images and 3D data generated during the SLAM process. It receives and processes data such as poses, RGB images, depth images, meshes, and point clouds from the SLAM system and displays them in real-time using Open3D\cite{zhou2018open3d} and Matplotlib\cite{hunter2007matplotlib}. Core functionalities include trajectory display, calculation and presentation of image quality metrics, and saving of visualization results. Immediate visualization effects enable researchers to quickly debug and optimize algorithms.

The \textit{OfflineViewer} is used to replay the results of SLAM algorithms. It reads trajectory data and updates images, point clouds, and meshes in the interface accordingly. Depending on the configuration, the \textit{OfflineViewer} can also save rendering results as videos.

\subsubsection{\textbf{Metrics and Evaluation}}

XRDSLAM supports obtaining and saving the poses of tracking frames, rendered images, as well as point clouds or meshes of the scene, and calculates related evaluation metrics based on ground truth data. The \textit{EvalMetrics} class provides an evaluation process for the results of SLAM. The overall evaluation of the SLAM system covers three main aspects: rendering quality, 3D reconstruction quality, and trajectory accuracy, with specific metrics including: 

{\textbf{Trajectory Accuracy:}} In the field of SLAM, \textit{ATE}\cite{sturm2012benchmark} is a widely used accuracy metric, and we adopt this metric to evaluate the trajectory accuracy.

{\textbf{Rendering Metrics:}} We adopt \textit{PSNR} (Peak Signal to Noise Ratio) \cite{hore2010image}, \textit{SSIM} (Structural Similarity Index Measure)\cite{hore2010image}, and \textit{LPIPS} (Learned Perceptual Image Patch Similarity)\cite{zhang2018unreasonable} to evaluate the quality of novel view rendering.

{\textbf{Reconstruction Quality:}} 
Similar to the paper\cite{tosi2024nerfs,zhu2022nice}, we use the commonly used 7 metrics to comprehensively assess the quality of the reconstruction, which include
\textit{Accuracy}\cite{sucar2021imap}, \textit{Completion}\cite{sucar2021imap}, \textit{Completion Ratio}\cite{sucar2021imap}, \textit{Precision}\cite{knapitsch2017tanks}, \textit{Recall}\cite{knapitsch2017tanks}, \textit{F-Score}\cite{knapitsch2017tanks}, \textit{L1-Depth}\cite{zhu2022nice}.


\begin{table*}[h]
    \centering
    \caption{We evaluate the integrated methods based on metrics in 3 aspects: \textit{Trajectory}, \textit{Rendering}, and \textit{Reconstruction}. The superscript $^1$ means the metrics are from the original papers and \cite{tosi2024nerfs}. '-' means not supported or not provided in the paper. '$^*$' means the version in XRDSLAM. Bolded fonts indicate better metrics.}
    \begin{tabular}{cccccccccccc}
            \toprule
            \multirow{2}{*}{\textbf{Algorithm}} & \textbf{Trajectory}& \multicolumn{3}{c}{\textbf{Rendering}}&  \multicolumn{7}{c}{\textbf{Reconstruction}} \\
        &  ATE(cm)$\downarrow$ &  PSNR$\uparrow$ &  SSIM$\uparrow$&  LPIPS$\downarrow$&  \makecell[c]{Precision\\(\%)$\uparrow$} &  \makecell[c]{Recall\\(\%)$\uparrow$} &  F1(\%)$\uparrow$ &  \makecell[c]{Depth L1\\(cm)$\downarrow$}
 & \makecell[c]{Acc. \\ (cm) $\downarrow$}& \makecell[c]{Comp. \\ (cm)$\downarrow$}& \makecell[c]{Comp. Ratio \\ \textless 5cm (\%) $\uparrow$} \\
        \midrule
         NICE-SLAM\footnotemark[1]&  \textbf{1.95}&  24.42&  0.81&  \textbf{0.23}&  44.10&  \textbf{43.69}&  \textbf{43.86}&  3.53
 & 2.85& \textbf{3.00}&\textbf{89.33} \\
         NICE-SLAM$^*$&  2.09&  \textbf{25.68}&  \textbf{0.85}&  0.32&  \textbf{46.62}&  37.53&  41.47&  \textbf{2.62}
 & \textbf{2.03}& 3.38&87.81 \\
        \hdashline[0.5pt/3pt]
         Co-SLAM\footnotemark[1]&  \textbf{0.86}&  30.24&  0.93&  0.25&  -&  -&  -&  \textbf{1.51}
 & 2.10& \textbf{2.08}&\textbf{93.44} \\
         Co-SLAM$^*$&  1.11&  \textbf{30.34}&  0.93&  \textbf{0.24}&  80.66&  68.79&  74.23&  1.63
 & \textbf{1.53}& 2.90&89.81 \\
        \hdashline[0.5pt/3pt]
         Vox-Fusion\footnotemark[1]&  \textbf{0.54}&  24.41&  0.8&  \textbf{0.24}&  55.73&  49.13&  52.2&  2.46
 & 2.37& \textbf{2.28}&\textbf{92.86} \\
         Vox-Fusion$^*$&  0.56&  \textbf{27.95}&  \textbf{0.90}&  0.25&  \textbf{89.52}&  \textbf{71.34}&  \textbf{79.39}&  \textbf{1.03}
 & \textbf{1.39}& 2.82&90.13 \\
        \hdashline[0.5pt/3pt]
         Point-SLAM\footnotemark[1]&  0.52&  \textbf{35.17}&  0.97&  0.12&  96.99&  83.59&  89.77&  0.44
 & -& -&- \\
         Point-SLAM$^*$&  \textbf{0.47}&  34.1&  0.97&  \textbf{0.10}&  \textbf{99.3}&  \textbf{83.78}&  \textbf{90.86}&  \textbf{0.38}
 & 1.25& 3.12&88.15 \\
        \hdashline[0.5pt/3pt]
         SplaTAM\footnotemark[1]&  \textbf{0.36}&  34.11&  0.97&  0.1&  -&  -&  -&  -
 & -& -&- \\
         SplaTAM$^*$& 0.4& \textbf{34.44}& \textbf{0.96}& \textbf{0.09}& -& -& -& -
 & -& -&- \\
        \hline
         DPVO$^*$& 0.31& -& -& -& -& -& -& -
 & -& -&- \\
         NeuralRecon$^*$& -& -& -& -& 13.29& 7.43& 9.51& - & 5.87& 19.36&38.13 \\
         \bottomrule
    \end{tabular}
    
    \label{tab:SLAM_Evaluation}
\end{table*}


\section{Experiment}
\subsection{SLAM Implementations}
Table \ref{tab:SLAM_implements} shows screenshots of various algorithms running in a 3D visualizer. These include 5 SLAM methods, an odometry method, and an incremental reconstruction method, all integrated into XRDSLAM.

Additionally, XRDSLAM can flexibly support various scene encoding formats and mapping methods, such as NICE-SLAM's multi-resolution feature grids, Co-SLAM's hash grid, Vox-Fusion's sparse voxel octree, Point-SLAM's neural point cloud, and SplaTAM's 3DGS.

\subsection{Test Datasets}
To facilitate the comparative evaluation of existing SLAM algorithms, we chose the Replica\cite{straub2019replica} dataset as a benchmark. The Replica dataset provides real, high-quality spatial reconstruction results, ground truth trajectories, and RGBD data, meeting the needs for testing and evaluating most current SLAM algorithms. We used 8 indoor RGBD sequences for evaluation. Additionally, developers can refer to the configuration parameters provided in the original paper for comparative evaluations on other datasets.

\subsection{Evaluation}
\subsubsection{\textbf{Accuracy Evaluation}}

XRDSLAM provides an evaluation module for detailed assessment. We compare the accuracy of algorithms using the original resolution of the \textit{Replica} dataset and the default parameters provided by the original algorithm repository. Each \textit{Replica} dataset contains 2000 frames of images, and we perform scene rendering every 50 frames to obtain RGB and depth images. The \textit{Tracker} process updates the trajectory estimation in real-time and automatically saves the 3D mesh after the operation is completed. During the evaluation process, DPVO requires the scale correction because DPVO uses only RGB data without real scale.

Table \ref{tab:SLAM_Evaluation} shows the average results of metrics on 8 Replica datasets for multiple original algorithms and their integrated versions (with '$^*$') in XRDSLAM. The better results between the original ones and XRDSLAM-integrated ones are highlighted in bold. The complete evaluation table can be viewed on our github repository. 

From the Table\ref{tab:SLAM_Evaluation}, it's obvious that the algorithms integrated into XRDSLAM are comparable to their original versions in terms of accuracy. In terms of 3D reconstruction quality, Point-SLAM performs the best in the \textit{Accuracy} metric. In terms of image rendering quality, both Point-SLAM and SplaTAM show excellent performance. In terms of localization metrics, DPVO has the highest accuracy, and SplaTAM's localization accuracy is superior to other neural mapping-based methods.

\begin{figure}[h]
    \centering
    \begin{minipage}{0.23\textwidth}
        \centering
        \includegraphics[width=\linewidth]{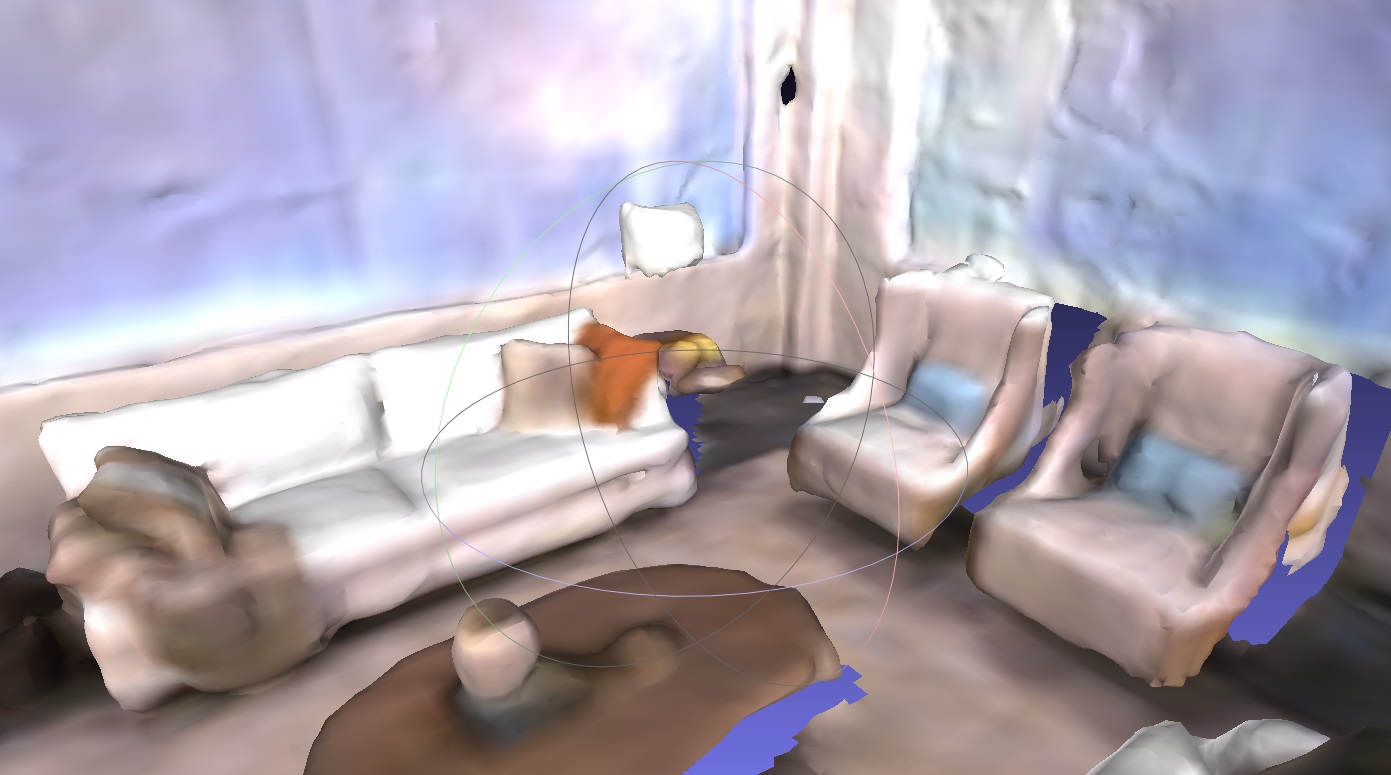}
        \caption*{NICE-SLAM}  
    \end{minipage}
    \hfill
    \begin{minipage}{0.23\textwidth}
        \centering
        \includegraphics[width=\linewidth]{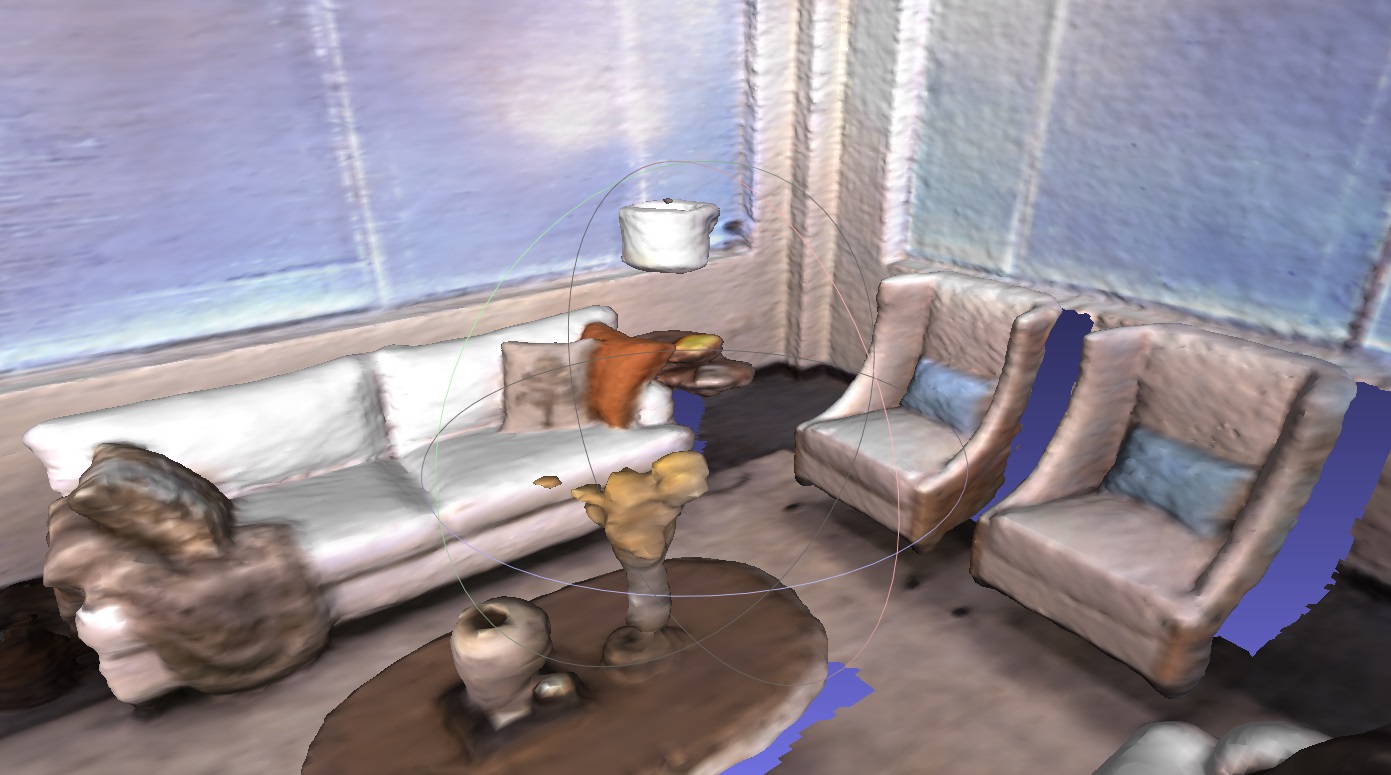}
        \caption*{Co-SLAM}
    \end{minipage}
    \hfill
    \begin{minipage}{0.23\textwidth}
        \centering
        \includegraphics[width=\linewidth]{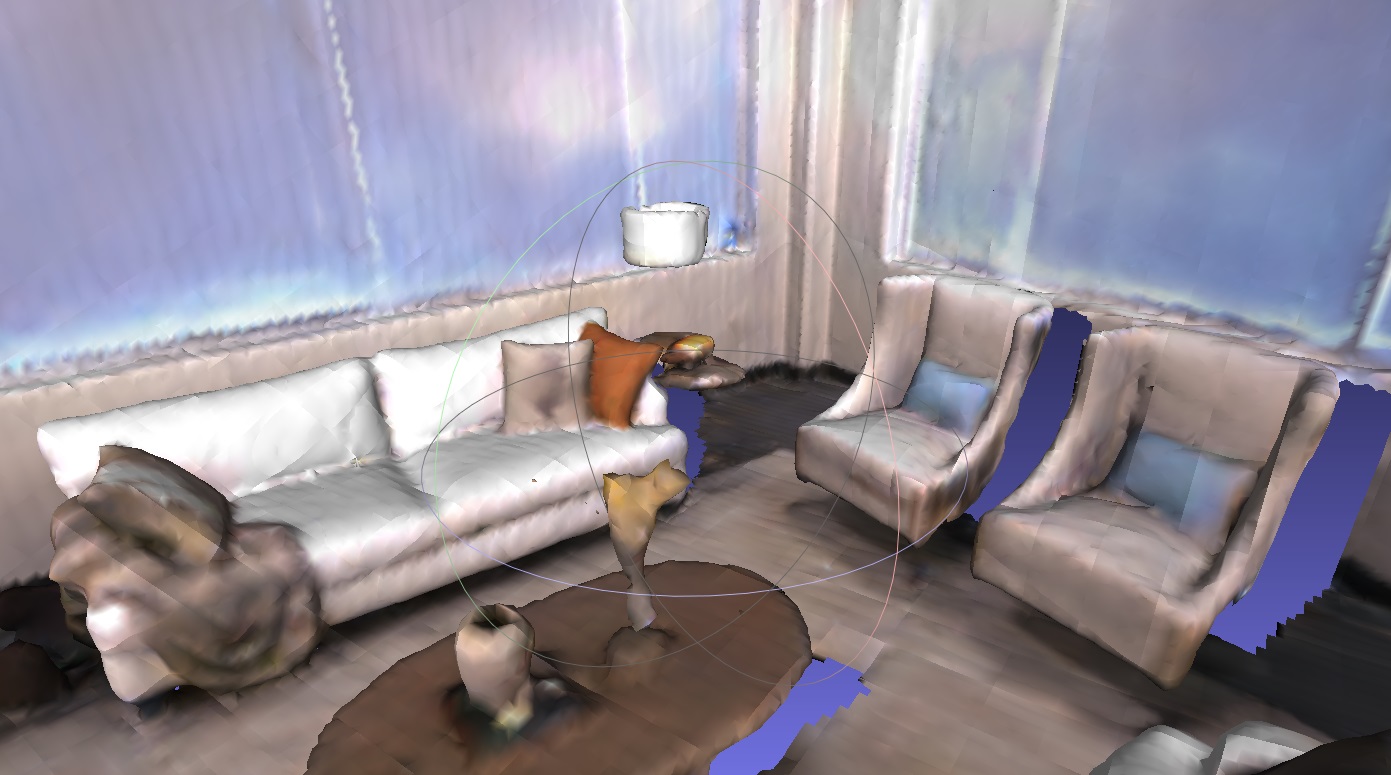}
        \caption*{Vox-Fusion}
    \end{minipage}
    \hfill
    \begin{minipage}{0.23\textwidth}
        \centering
        \includegraphics[width=\linewidth]{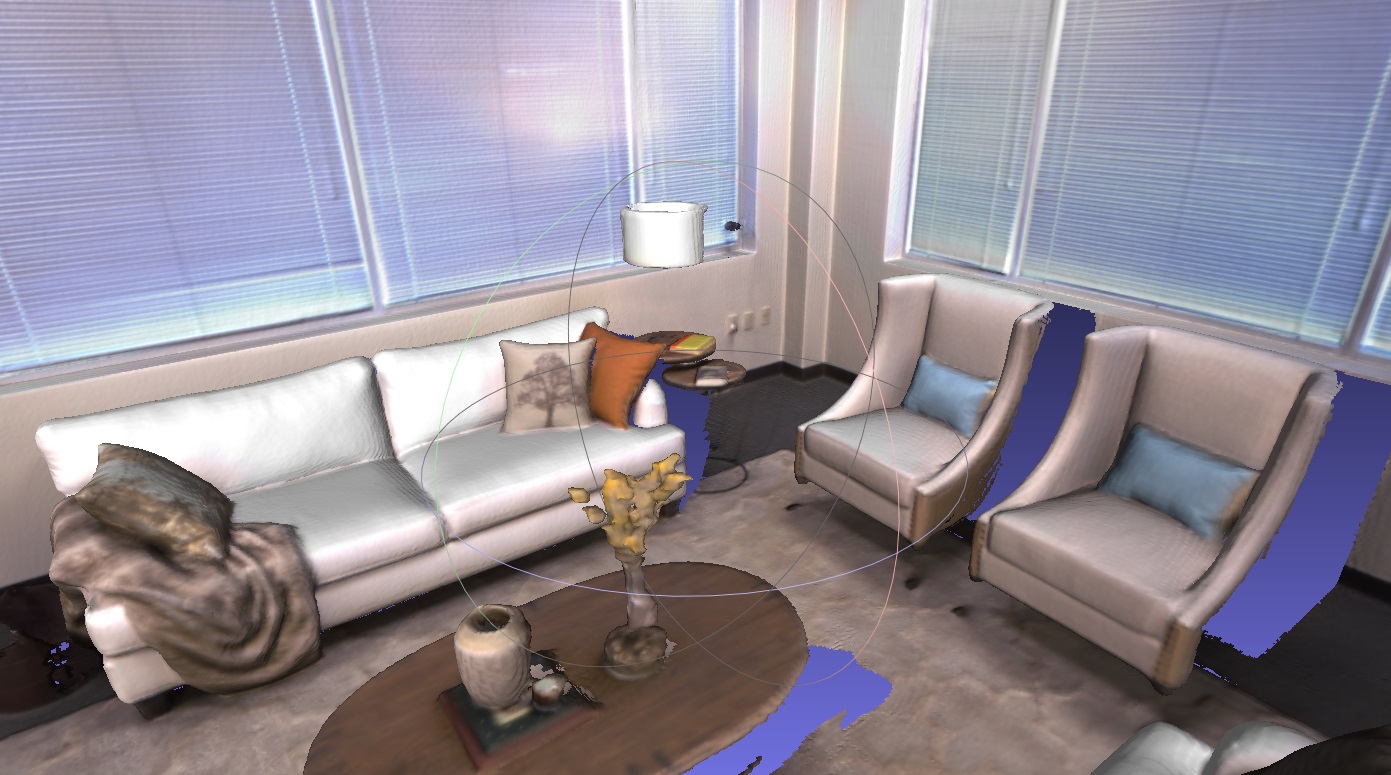}
        \caption*{Point-SLAM}
    \end{minipage}
    
    \caption{Comparison of mesh reconstruction of different integrated algorithms on dataset \textit{Replica}.}
    \label{fig:mesh_eval}
\end{figure}

Fig.\ref{fig:mesh_eval} presents the qualitative results of several SLAM methods integrated into XRDSLAM for mesh reconstruction on the Replica/room0 dataset, with performance progressively improving from NICE-SLAM to Co-SLAM, Vox-Fusion, and finally, Point-SLAM.

\subsubsection{\textbf{Performance Evaluation}}

We use Python's tqdm library to tally the running time per frame and calculate the FPS for each algorithm. Table \ref{tab:SLAM_implements} summarizes our benchmark results of running XRDSLAM on the downsampled Replica/office0 dataset (resolution of 600$\times$340). The tests are conducted on an NVIDIA GTX 1080Ti graphics card (with 12GB of memory). Since the NeuralRecon only supports a resolution of 640x480, images are cropped and resized from the original to this resolution for testing.

From Table \ref{tab:SLAM_implements}, among the algorithms integrated into XRDSLAM, NeuralRecon has the highest efficiency. Among the SLAM algorithms, Co-SLAM using hash grids performs the best in terms of efficiency, while Point-SLAM and SplatTAM are relatively less efficient. Although SplatTAM is more efficient in image rendering, the overall system efficiency is not ideal due to the need to process each frame and the higher number of iterations.

Additionally, the FPS of NICE-SLAM within XRDSLAM is very different from the results in the paper\cite{tosi2024nerfs}, primarily because the original NICE-SLAM defaults to extracting a mesh during operation, leading to longer time consumption. XRDSLAM standardizes the data saving process, making the comparison of algorithm performance more equitable. Factors affecting algorithm efficiency include not only data saving and the algorithm's own process but also the number of iterations, the number of sampling points, and the data processing interval frames. To ensure consistency with the original algorithm's accuracy, the default configuration of the original algorithm was used during the evaluation phase.

Table \ref{tab:SLAM_implements} also provides the peak GPU memory usage (in GB) of XRDSLAM when running on the Replica/office0 dataset (with a downsampled resolution of 600x340). During the tests, Vox-Fusion's GPU memory initially spiked to around 10GB, then dropped and stabilized between 4-6GB. SplaTAM's memory usage is closely tied to the resolution, primarily because it uses RGB-D point clouds for GS initialization. Point-SLAM, which is based on neural point clouds, sees increasing memory consumption as the point cloud map expands. CoSLAM and NICE-SLAM, which are based on fixed-resolution voxel grids and a constant number of samples per frame, have relatively stable memory usage. NeuralRecon, being an incremental reconstruction method, also experiences increasing memory consumption as the map grows.

\section{Conclusion}

This article introduces a new general SLAM framework—XRDSLAM. The framework provides a common interface and tools, making the platform easy to use. The modular design facilitates the integration and migration of different algorithms, while providing unified result export and evaluation functions, achieving a fair and convenient comparison of different SLAM algorithms. XRDSLAM significantly reduces the cost of code development and improves development efficiency, promoting the development of SLAM technology.

On this platform, we integrate the state-of-the-art SLAM algorithms (Vox-Fusion, NICE-SLAM, SplaTAM and etc.) and conducted objective evaluations under a unified framework, scientifically analyzing the accuracy characteristics and computational efficiency of different algorithms. 

In the future, we will continue to integrate more of the most advanced SLAM algorithms, further perfecting the ecosystem of deep learning SLAM, and we welcome more community efforts to develop on this platform and contribute high-quality algorithms and code.








\bibliographystyle{plain}
\bibliography{reference}

\end{document}